\documentclass[letterpaper]{article} 
\usepackage{aaai23}  
\usepackage{times}  
\usepackage{helvet}  
\usepackage{courier}  
\usepackage[hyphens]{url}  
\usepackage{graphicx} 
\usepackage{natbib}  
\usepackage{caption} 
\usepackage{newfloat}
\usepackage{listings}

\usepackage{algorithm}
\usepackage{latexsym}
\usepackage{multirow}
\usepackage{soul}
\usepackage{amsmath}
\usepackage{amssymb}
\usepackage{makecell}
\usepackage[noend]{algpseudocode}
\usepackage{booktabs}
\usepackage[utf8]{inputenc}
\usepackage{microtype}
\usepackage{adjustbox}
\DeclareCaptionStyle{ruled}{labelfont=normalfont,labelsep=colon,strut=off} 
\floatstyle{ruled}
\newfloat{listing}{tb}{lst}{}
\floatname{listing}{Listing}
\title{Effective Open Intent Classification with K-center Contrastive Learning and Adjustable Decision Boundary}
\author{
    Xiaokang Liu\textsuperscript{\rm 1}\equalcontrib,
    Jianquan Li\textsuperscript{\rm 2}\equalcontrib,
    Jingjing Mu\textsuperscript{\rm 2},
    Min Yang\textsuperscript{\rm 3}\thanks{Min Yang is corresponding author.},
    Ruifeng Xu\textsuperscript{\rm 4},
    Benyou Wang\textsuperscript{\rm 5,6}
}
\affiliations {
        \textsuperscript{\rm 1} China Automotive Technology and Research Center Co., Ltd. \\
    \textsuperscript{\rm 2} Beijing Ultrapower Software Co.,Ltd. \quad \\
    \textsuperscript{\rm 3} Shenzhen Institutes of Advanced Technology, Chinese Academy of Sciences \\
     \textsuperscript{\rm 4} Harbin Institute of Technology, Shenzhen \\
    \textsuperscript{\rm 5} The Chinese University of Hong Kong (Shenzhen)\\
     \textsuperscript{\rm 6} Shenzhen Research Institute of Big Data, The Chinese University of Hong Kong, Shenzhen, China \\
    liuxiaokang1@catarc.ac.cn, \{lijianquan2,mujingjing\}@ultrapower.com.cn, min.yang@siat.ac.cn,  xuruifeng@hit.edu.cn,wangbenyou@cuhk.edu.cn
  }

\usepackage{bibentry}

\begin{document}

\maketitle

\begin{abstract}
Open intent classification, which aims to correctly classify the known intents into their corresponding classes while identifying the new unknown (open) intents, is an essential but challenging task in dialogue systems. 
In this paper, we introduce novel K-center \textbf{\underline{c}}ontrastive \textbf{\underline{l}}earning and \textbf{\underline{a}}djustable decision \textbf{\underline{b}}oundary learning (CLAB) to improve the effectiveness of open intent classification. First, we pre-train a feature encoder on the labeled training instances, which transfers knowledge from known intents to unknown intents.
Specifically, we devise a K-center contrastive learning algorithm to learn discriminative and balanced intent features, improving the generalization of the model for recognizing open intents. 
Second, we devise an adjustable decision boundary learning method with expanding and shrinking (ADBES) to determine the suitable decision conditions. Concretely, we learn a decision boundary for each known intent class, which consists of a decision center and the radius of the decision boundary.
We then expand the radius of the decision boundary to accommodate more in-class instances if the out-of-class instances are far from the decision boundary; otherwise, we shrink the radius of the decision boundary. 
Extensive experiments on three benchmark datasets clearly demonstrate the effectiveness of our method for open intent classification. 
For reproducibility, we submit the code at: https://github.com/lxk00/CLAP
\end{abstract}

\section{Introduction}
Intent classification is an essential task in dialogue systems by categorizing input sequences into pre-defined intent classes. 
It is usually formulated as a supervised classification problem. Recently, superior classification performance has been achieved by deep neural networks \cite{bendale2016towards,hendrycks2016baseline}, where sufficient labeled training data is provided during model training. However, these methods make a closed world assumption which is impractical for real-world applications, and cannot deal with the previously unseen intent classes \cite{2017Neural,2020Out}. 

Open intent classification, which aims to correctly classify the known intents into their corresponding classes while identifying the unknown intents as the ``open intent'', improves customer satisfaction by reducing false-positive errors, making the intent classification more challenging. 
Early studies employ SVM to reject unseen class examples~\cite{2004Support,2004In,2014Multi} based on feature engineering which is labor-intensive. 
Recently, there are several works being proposed to detect open intents by learning a decision boundary of each known intent in the similarity space \cite{bendale2016towards,hendrycks2016baseline,shu2017doc}. Among them, ADB~\cite{zhang2021deep} is a representative open intent classification method, which learns the adaptive spherical decision boundary for each known class with the aid of a pre-trained text encoder. 

Despite the noticeable progress of previous works, there are still several challenges for open intent classification, which are not addressed well in prior studies. First, the superior performance of standard supervised learning is highly dependent on the quality and balance of the training labels. The noisy and unbalanced training data may significantly degrade the generalization performance of deep models on out-of-distribution intents. 
Although KCL \cite{kang2020exploring} combined the strengths of supervised learning and contrastive learning to learn representations that were both discriminative and balanced, it merely shortens the distance between the anchor sample and the other positive samples, but ignores the distances among the chosen positive samples.
To address this limitation, we learn more robust representations by incorporating instance-level semantic discriminativeness into representation learning via contrastive learning. Concretely, we propose a K-center contrastive learning (KCCL) method to learn more robust and balanced text representations. Different from the general supervised contrastive loss, 
our KCCL method encourages each positive instance to move in the direction of the other $K$ positive instances so that each positive instance moves towards the center of the positive samples, resulting in better clustering performance and speeding up the convergence. In this way, we can better gather the semantic features with the same intent, resulting in better representation space for learning the intent decision boundary.

Second, ADB~\cite{zhang2021deep} is a representative method to learn the adaptive decision boundary for open intent classification, which learns the radius of a specific class boundary by solely considering the instances belonging to the given class.
In particular, the ADB method only takes known intents to adjust the boundary and disregards negative samples, which can lead to performance deterioration on out-of-distribution intents. 
For example, when plenty of unknown intents fall inside the decision boundary, the ADB method cannot make correct adjustments (i.e., shrink boundary) accordingly. Since the goal of open intent classification is to distinguish the known intents from the unknown ones, the decision boundary learning should consider both known and unknown intents. 
To solve the above problem and learn a more accurate decision boundary of each intent class, we devise an expanding and shrinking approach to further adjust the decision boundary by considering out-of-class (negative) instances for determining the intent decision boundary. Specifically, we can expand the radius of the decision boundary to accommodate more in-class instances if out-of-class instances are far from the decision boundary; otherwise, we shrink the radius of the decision boundary. 

Our contributions are summarized as follows:
\begin{itemize}
    \item We propose a K-center contrastive learning method to learn robust text representations and make a clear margin of different intents by gathering the semantic features with the same intent. 
    \item We propose an adjustable spherical decision boundary learning method to determine accurate and suitable decision conditions of each intent class by expanding or shrinking the radius of the decision boundary.  
    \item We conduct extensive experiments on three benchmark datasets. Experimental results show that our method achieves better performance than the state-of-the-art methods for open intent classification.
\end{itemize}

\section{Related Work}
\subsection{Open Intent Classification}
Open intent classification, which aims to correctly classify the known intents into their corresponding classes while identifying the new unknown intents, has attracted noticeable attention due to its broad applications in dialogue systems and question answering. Early approaches leverage machine learining methods (e.g., SVM) to reject unseen classes for open world recognition~\cite{2001Estimating, 2004Support, 2004In, 2014Multi}. However, these methods relied heavily on feature engineering which is labor intensive. 

With the advance of deep learning, deep neural networks have dominated the literature of open intent classification, which can learn the deep semantic features of input texts automatically. \citeauthor{brychcin2016unsupervised} \shortcite{brychcin2016unsupervised} calibrated the confidence of the softmax outputs to compute the calibrated confidence score and leveraged the score to obtain the decision boundary for unknown intent detection.  \citeauthor{yu2017open} \shortcite{yu2017open} leveraged adversarial learning to produce positive and negative samples from known samples for training the open intent classifier.  \citeauthor{ryu2018out} \shortcite{ryu2018out} exploited the generative adversarial networks \cite{goodfellow2014generative} to generate out-of-distribution (OOD) samples with the generator and learned to reject produced OOD samples with the discriminator. However, these methods need a confidence score, which is difficult to be pre-defined in advance, to determine the likelihood of an utterance being out-of-scope.

Subsequently, \citeauthor{lin2019deep} \shortcite{lin2019deep} proposed the DeepUNK method, which utilized a margin-based method to train the classifier and detected the unknown intent with local outlier factor. 
\citeauthor{zhang2021deep} \shortcite{zhang2021deep} introduced a post-processing method to learn a spherical decision boundary for each known intent class with the aid of a pre-trained intent encoder. 
\citeauthor{prem2021unknown} \shortcite{prem2021unknown} presented a post-processing method, which utilized  multi-objective optimization to tune a deep neural network based intent classifier and  make the classifier capable of detecting unknown intents. 
\citeauthor{zhan2021out} \shortcite{zhan2021out} constructed a set of pseudo outliers by using inliner features via self-supervision in the training stage. The pseudo
outliers were used to train a discriminative classifier that can be directly applied to and generalize well on the test task. However, these approaches do not take full advantage of the representation learning for learning better discriminative and balanced features.

\subsection{Contrastive Learning} 
Self-supervised learning (SSL) has attracted much attention since it can avoid manually annotating large-scale datasets and uses the learned representation well for downstream tasks~\cite{Hendrycks2019UsingSL, Misra2020SelfSupervisedLO}. Contrastive learning is a popular self-supervised representation learning method~\cite{le2020contrastive}. The key idea behind contrastive learning is to narrow the distance between the anchor sample and the positive samples and increase the distance between the anchor sample and the negative samples. 
SimCSE~\cite{gao2021simcse} is a representative contrastive learning method, which learned superior sentence representations from either unlabeled or labeled data.  KCL~\cite{kang2020exploring} effectively combined the strengths of the supervised method and the contrastive learning method to learn representations that were both discriminative and balanced. CERT~\cite{fang2020cert} created augmentation samples of sentences using back-translation, and then fine-tuned a pretrained language representation model by predicting whether two augmentation samples are from the same original sentence or not. 

\begin{figure*}
    \centering
    \includegraphics[height=170pt]{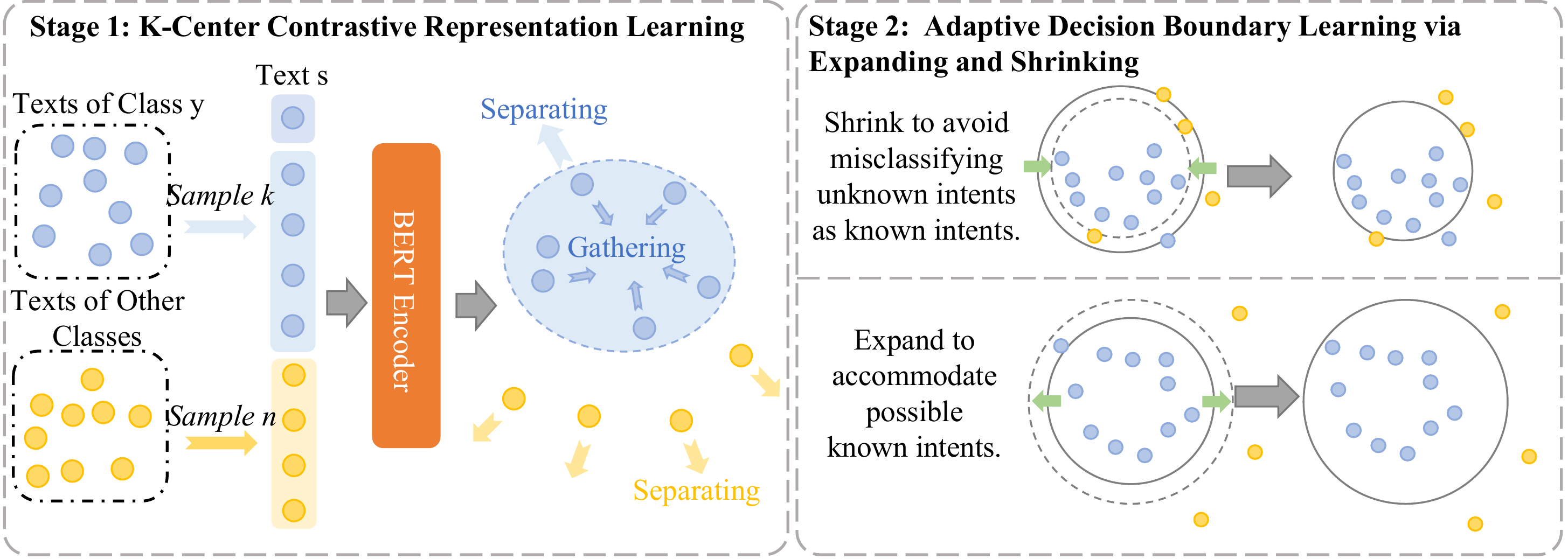}
    \caption{An overview of our method. In the first stage (left), we leverage a novel K-center contrastive learning method to improve the text representation learning.  In the second stage (right), we design the adjustable decision boundary learning with expanding and shrinking to determine the suitable decision conditions for the known intents.} 
    \label{fig:model}
\end{figure*}

\section{Our Methodology}
\paragraph{Problem Definition}
Suppose we have a training data set $\mathcal{S}=\{(x_i, y_i)\}_{i}^{N}$ of $N$ input instances, where each input sequence $x_i=\{w_1, \ldots,w_n\}$ has a corresponding output intent label $y_i\in \{1, \ldots, C\}$ and $n$ is the sequence length. Here, $C$ denotes the number of different intent classes in the training dataset $\mathcal{S}$. The goal of open intent classification is to  correctly classify the $K$ known intents into their corresponding classes while identifying the unknown intents as the ($C+1$)$th$  intent type (also called ``open intent'').   

\paragraph{Model Overview}
As illustrated in Figure \ref{fig:model}, our open intent classification method can be divided into two training stages. In the first stage, we employ K-center contrastive learning method to learn the sentence representations. In the second stage, we design a novel adaptive boundary learning method to comprehensively learn the in-distribution and out-of-distribution data, so that the decision boundary can be clearly determined.

\subsection{Initial Representation Learning}
Following previous work \cite{zhang2021deep}, we employ the pre-trained BERT~\cite{Devlin2019BERTPO} as sentence encoder to learn the feature representations of input sentences. In particular, for each input sentence $x_i$, we obtain the hidden representations of tokens $[CLS, \mathbf{v}_1, \mathbf{v}_2, \ldots, \mathbf{v}_n, SEP] \in \mathbb{R}^{(n+2)\times H}$ from the last layer of BERT, where $\mathbf{v}_i$ is the hidden representation of the $i$-th token and $H=768$ denotes the dimension size of token embeddings. $CLS$ and $SEP$ represent the beginning and ending symbols. We use mean-pooling over all tokens representations to obtain the sentence representations $\mathbf{o}_i$ for input sequence $x_i$:
\begin{equation}
\mathbf{o}_i = {\mathrm mean\textit{-}pooling}([CLS, \mathbf{v}_1, \ldots, \mathbf{v}_n, SEP])
\end{equation}

To further enhance the feature representation learning, we use the ReLU activation function following the linear mapping to get the feature representation $\mathbf{h}_i\in\mathbb{R}^H$, and then obtain $\mathbf{z}_i$ by normalizing $\mathbf{h}_i$:
\begin{gather}
    \mathbf{h}_i ={\rm ReLU}(W_1 \mathbf{o}_i+ \mathbf{b}_1), ~~\mathbf{z}_i = \frac{\mathbf{h}_i }{\left\| \mathbf{h}_i  \right\|_2}
\end{gather}
where $W_1$  and $\mathbf{b}_1$ are learnable parameters. $\left\| \cdot  \right\|_2$ are Euclidean normalization over the input vector.

\subsection{K-center Contrastive Representation Learning}
The general supervised contrastive learning~\cite{Oord2018RepresentationLW} aims to learn an embedding space where the instances from the same category are pulled closely and the instances from different category are pushed apart. Specifically, we minimize a contrastive loss function $\mathcal{L}_{\rm CL}$ to optimize the  encoder network:
\begin{equation}\label{eq:cl}
\begin{aligned}
\mathcal{L}_{\rm CL}= 
-\frac{1}{N}\sum_{i=1}^{N}\log\frac{{e}^{\mathbf{z}_{i} \cdot \mathbf{z}_{i}^+/\tau}}{{e}^{\mathbf{z}_{i} \cdot \mathbf{z}_{i}^+/\tau} + \sum_{\mathbf{z}_i^- \in Z_i^-} {e}^{\mathbf{z}_{i} \cdot \mathbf{z}_{i}^-/\tau}}
\end{aligned}
\end{equation}
where $\mathbf{z}_i^+$ indicates one positive sample randomly chosen from the training set which has the same intent label with $\mathbf{z}_i$. $\pmb{Z}_i^-$ indicates the set of randomly chosen negative training samples which have different intent labels with $\mathbf{z}_i$. $\tau$ is a temperature value. 

The above supervised contrastive loss defined in Eq. (3) learns more robust representations by incorporating instance-level semantic discriminativeness into the representation learning. However, conventional supervised contrastive learning uses all the samples from the same class to construct the positive pairs, which could result in the dominance of instance-rich classes in the representation learning, as revealed in \cite{kang2020exploring}. 

\paragraph{\textbf{K-center Contrastive Learning}}
To alleviate the problem of dominating the instance-rich classes, we propose a K-center contrastive learning (KCCL) method to learn more balanced feature representations, inspired by \cite{kang2020exploring}. Different from \cite{kang2020exploring} that compares $K$ positive instances and the negative instances, we also compare each instance pair within the set of $K$ positive instances. To be specific, for each training instance $\mathbf{z}_i$ with intent label $y_i$, we randomly select $K$ samples with intent class $y_i$ to form the positive sample set $\pmb{Z}^+_i$. We also randomly select $M$ training samples from other intent classes as the negative sample set $\pmb{Z}^-_i$. The K-center contrastive loss function $\mathcal{L}_{\rm KCCL}$ can be formalized as follows:
\begin{equation}
\small
\begin{aligned}
&\mathcal{L}_{\rm KCCL}=-\frac{1}{N \cdot K\cdot(K+1)}\sum_{i=1}^{N}\sum_{m=1}^{K+1}\sum_{n=1, n \neq m}^{K+1} \\
& \log\frac{e^{\mathbf{z}_{i,m}^{+} \cdot \mathbf{z}_{i,n}^{+}/\tau}}{e^{\mathbf{z}_{i,m}^{+}\cdot \mathbf{z}_{i,n}^{+}/\tau}+\sum_{\pmb{z}_i^- \in \pmb{Z}_{i}^-}{(e^{\mathbf{z}_{i,m}^{+}\cdot \pmb{z}_i^-/\tau} + e^{\mathbf{z}_{i,n}^{+}\cdot \mathbf{z}_{i}^{-}/\tau}})}
\end{aligned}
\end{equation}
where $\mathbf{z}_{i,m}^{+}$ and  $\mathbf{z}_{i,n}^{+}$ are two positive samples in $\pmb{Z}^+_i$. Different from the previous contrastive loss defined in Eq. (\ref{eq:cl}), our KCCL method additionally shortens the distances between positive samples. In this way, we can better gather the semantic features with the same intent \cite{Oord2018RepresentationLW,kang2020exploring},  resulting in better representation space for learning the intent decision boundary. 

In addition to supervised contrastive loss, we also leverage supervised cross-entropy (CE) loss $\mathcal{L}_{\rm CE}$ to learn semantic discriminative features: 
\begin{equation}
    \label{eq:crossentropyloss}
    \mathcal{L}_{\rm CE} =-\frac{1}{N}\sum_{i=1}^{N} \mathbf{y}_{i}\log {\rm softmax}(W_2 \mathbf{z}_i+\mathbf{b}_2)
\end{equation}
where $\mathbf{y}_{i}$ is the one-hot vector for intent label $y_i$. $W_2$  and $\mathbf{b}_2$ are learnable parameters. 

Overall, the loss function $\mathcal{L}_{S_1}$ of the first stage training is obtained as a weighted sum between  $\mathcal{L}_{\rm KCCL}$ and  $\mathcal{L}_{\rm CE}$:
\begin{equation}
    \mathcal{L}_{S_1} =  \lambda \cdot \mathcal{L}_{\rm KCCL}  +(1-\lambda)\cdot \mathcal{L}_{\rm CE} 
\end{equation}
where $\lambda$ is a pre-defined hyperparameter for controlling the impact of the two loss functions.

\subsection{Adaptive Decision Boundary Learning via Expanding and Shrinking}
In this section, we propose a novel inflating and shrinking approach to learn adaptive decision boundaries for open intent classification.

\subsubsection{Decision Boundary Learning}
After learning the representations of input sentences via the KCCL method, we learn a decision boundary for each known intent class $k$, which consists of a decision center $\mathbf{c}_{k}$ and the radius of the decision boundary $\mathrm{\Delta}_{k}$. Formally, suppose there is a dataset $\mathcal{S}_k$ containing all the training instances with intent class $k$. The decision center $\mathbf{c}_{k}$ is a fixed vector and the radius of the decision boundary $\mathrm{\Delta}_{k}$ is a learnable parameter. The decision center $\mathbf{c}_{k}$ of class $k$ can be acquired by averaging the representations of instances in $\mathcal{S}_k$: 
\begin{equation}
    \mathbf{c}_k=\frac{1}{|\mathcal{S}_k|}\sum_{\mathbf{z}_i\in \mathcal{S}_k} \mathbf{z}_i
\end{equation}
where $|\mathcal{S}_k|$ denotes the number of instances in $\mathcal{S}_k$.
For each known instance $\mathbf{z}_i$ with the latent class $k$, the radius $\mathrm{\Delta}_{k}$ of the decision boundary should satisfy the following constraints:
\begin{equation}
\left \| \mathbf{z}_i - \mathbf{c}_{k} \right \|_2 \leq \Delta_{k}
\end{equation}

However, it is non-trivial to optimize the radius $\mathrm{\Delta}_{k}$ of the decision boundary for open intent classification. As revealed in \cite{zhang2021deep}, the radius should be large enough to surround known intent samples as much as possible so as to reduce the empirical risk, while we should also decrease the open space risk by shrinking the radius to avoid 
To balance the trade-off between the open space risk and the empirical risk, the adaptive decision boundary (ADB) method \cite{zhang2021deep} is proposed to optimize each learnable boundary radius as follows:
\begin{equation}
\label{eq:ADB}
    \begin{aligned}
    \mathcal{L}_{\rm ADB}= \frac{1}{N}\sum_{i=1}^{N}\gamma_i \cdot (\left \| \mathbf{z}_i - \mathbf{c}_{y_i}\right \|_2-\Delta_{y_i}) + \\ (1 - \gamma_i) \cdot (\Delta_{y_i} - \left \| \mathbf{z}_i - \mathbf{c}_{y_i}\right \|_2)
    \end{aligned}
\end{equation}
where $\mathbf{z}_i$ is the representation of input sequence $x_i$ obtained in the first stage. $y_i$ is the label of $\mathbf{z}_i$. $\gamma$ is a variable to control the impact of the two constrains, which is defined as:
\begin{equation}
    \gamma_i =\begin{cases}
    1, & \text{ if } \left \| \mathbf{z}_i  -\mathbf{c}_{y_i} \right \|_2> \Delta_{y_i}, \\ 
    0, & \text{ if } \left \| \mathbf{z}_i  - \mathbf{c}_{y_i} \right \|_2\leq \Delta_{y_i}.
    \end{cases}
\end{equation}
The parameters can be updated via stochastic gradient descent (SGD).
As shown in Eq. (\ref{eq:ADB}), the ADB method learns the  radius of a specific class boundary by solely considering the instances belonging to the given class.

\subsubsection{Expanding and Shrinking}
To learn a more accurate and suitable decision boundary of each intent class, we devise an expanding and shrinking approach to further adjust the decision boundary by considering out-of-class (negative) instances for determining the intent decision boundary. Specifically, we can expand the radius of the decision boundary to accommodate more in-class instances if out-of-class instances are far from the decision boundary; otherwise, we shrink the radius of the decision boundary. According to our statistics, the distances from the unknown intents to the decision center conforms to a skewed normal distribution. We adjust the decision boundary radius according to both tails of the distribution. If there are large volumes of samples that fall in the right tail, then we can expand the decision boundary, and If there are large volumes of samples that fall in the left tail, then we should shrink the decision boundary.

Formally, we introduce an expansion parameter $e$ and a shrink parameter $s$ for better expanding and shrinking the decision boundary. For each training instance $\mathbf{z}_i$, if the distance between the out-of-class (negative) sample $\mathbf{z}_i^{-}$ and the decision center $\mathbf{c}_{y_i}$ is less than $\mathrm{\Delta}_{y_i}+s$, then we should reduce the value of the radius $\mathrm{\Delta}_{y_i}$. On the contrary, if the distance between the negative sample $\mathbf{z}_i^{-}$ and the decision center $\mathbf{c}_{y_i}$ is greater than $\mathrm{\Delta}_{y_i}+e$, then we increase the value of the radius $\mathrm{\Delta}_{y_i}$. 
We define a novel loss function $\mathcal{L}_{\rm ADBES}$ for adaptive decision boundary learning with expanding and shrinking operations (ADBES) as follows:
\begin{equation}
\small
    \begin{aligned}
    & \mathcal{L}_{\rm ADBES}  = \mathcal{L}_{\rm ADB} + 
    \frac{1}{N}\sum_{i=1}^{N} \bigg\{ \eta \cdot \alpha_i \cdot \Big[ \left\|\mathbf{z}_i^{-} - \mathbf{c}_{y_i}\right\|_2 - \\
    & ~~~~(\Delta_{y_i} + e)\Big] + \eta \cdot  \beta_i \cdot \Big[(\Delta_{y_i} + s) - \left\|\mathbf{z}_i^{-} - \mathbf{c}_{y_i}\right\|_2\Big] \bigg\}
    \end{aligned}
\end{equation}
where $\eta$ is a pre-defined hyperparameter for controlling the impact of the expanding and shrinking approach. $\mathbf{z}_i^{-}$ is a randomly chosen negative sample which has different intent label with $\mathbf{z}_i$. $\alpha$ and $\beta$ determine the  expanding or shrinking operations over the decision boundaries, which are defined as follows: 
\begin{gather}
    \alpha_i =\begin{cases}
    1, & \left\|\mathbf{z}_i^{-} - \mathbf{c}_{y_i}\right\|_2 > \Delta_{y_i} + e, \\ 
    0, & \left\|\mathbf{z}_i^{-} - \mathbf{c}_{y_i}\right\|_2 \leq \Delta_{y_i} + e.
    \end{cases}\\
    \beta_i =\begin{cases}
    1, & \left\|\mathbf{z}_i^{-} - \mathbf{c}_{y_i}\right\|_2 < \Delta_{y_i} + s, \\ 
    0, & \left\|\mathbf{z}_i^{-} - \mathbf{c}_{y_i}\right\|_2 \geq \Delta_{y_i} + s.
    \end{cases}
\end{gather}

With the loss function $\mathcal{L}_{\rm ADBES}$, we can learn more accurate and suitable decision boundaries, which not only effectively surround most of the known intent instances but also identify the open intent instances. 

\subsection{Open Intent Classification during Inference}
During the inference (testing) phase, the input samples may come from both known and unknown intent classes. Given each test sample $x_j$, we obtain its feature representation $\mathbf{z}_j$ via the K-center contrastive learning. With the learned radius of the decision boundaries, the intent class $\hat{y}_j$ can be predicted by thresholding the Euclidean distance between  $\mathbf{z}_j$ and the decision center $\mathbf{c}_k$ (i.e., $d(\mathbf{z}_j,\mathbf{c}_k)$) w.r.t. the corresponding radius $\Delta_{k}$. That is, the model will mark the target instance as ``Unknown'' if it is outside any of the learned decision boundaries. 
Formally, we predict the open intent as follows: 
\begin{equation}
    \hat{y}_j =\begin{cases}
    {\rm Unknown},  {\rm if}~ d(\mathbf{z}_j, \mathbf{c}_k) > \Delta_{k}, k\in \mathcal{Y}; \\ 
    {\rm arg}~{\rm min}_{k\in \mathcal{Y}}d(\mathbf{z}_j, \mathbf{c}_k), ~ {\rm otherwise}.
    \end{cases}
\end{equation}
where $\mathcal{Y}=\{1, \ldots,K\}$ represents the label space of the known intents. 

\section{Experimental Setup}
\subsection{Datasets}
We conduct extensive experiments on three publicly available benchmark datasets. \textbf{BANKING}~\cite{casanueva2020efficient} is a fine-grained dataset in the banking domain, which contains 77 intents and 13,083 customer service queries.
\textbf{OOS}~\cite{larson2019evaluation} is a dataset for intent classification and out-of-scope prediction. It consists of 150 intents, 22,500 in-domain queries and 1,200 out-of-domain queries.
\textbf{StackOverflow}~\cite{xu2015short} is a dataset released originally in Kaggle.com. It contains 3,370,528 technical question titles. We use the processed dataset which has 20 different classes and 1,000 samples for each class.
We provide the detailed statistics of the datasets in Table \ref{tab:commands}.

\subsection{Evaluation Metrics}
Following previous work \cite{zhang2021deep}, we adopt the widely used accuracy and macro F1-score as the evaluation metrics for open intent classification. We treat all unknown classes as an open class, and calculate the overall accuracy (denoted as Acc) and macro F1-score (denoted as F1) over all intent classes. In addition, to better evaluate the capability of the classifiers for identifying known and unknown intents, we also compute the macro F1-score over known intents (denoted as Known) and unknown intents (denoted as Unknown), respectively. 


\subsection{Baseline Methods}
We compare our method with the five state-of-the-art open intent classification methods, including {\verb|MSP|~\cite{hendrycks2016baseline}} that adopts a softmax prediction for out-of-distribution detection, {\verb|DOC|~\cite{shu2017doc}} that uses deep learning to build a multi-class classifier and leverages Gaussian fitting to tight the decision boundary, {\verb|OpenMax|~\cite{bendale2016towards}} that applies the concept of meta-recognition to the activation patterns in the penultimate layer to reduce the risk of open space, {\verb|DeepUnk|~\cite{lin2019deep}} that replaces softmax loss with margin loss to learn discriminative deep features by forcing the network to maximize inter-class variance and minimize intra-class variance, {\verb|ADB|~\cite{zhang2021deep}} that presents a new post-processing method for open intention classification by learning the spherical decision boundary for each known class. 

\subsection{Implementation Details}
Following the same settings as in \cite{zhang2021deep}, we keep some classes in the training set as unknown and integrate them back during testing. In particular, we vary the number of known
classes in the training set with the proporation of 25\%, 50\%,
and 75\% classes, and adopt all intent classes for testing.
It is noteworthy that the unknown classes are only utilized during testing. To conduct a fair evaluation and ensure the stability of our model, we run our model ten times and report the average results over ten runs of experiments. 

In the first training stage, we build our model on top of the pre-trained BERT model (base-uncased) with 12-layer transformer and adopt most of its default hyperparameter settings. We freeze all the parameters of BERT except the last transformer layer to speed up the training process and avoid over-fitting. The number of positive samples in KCCL are in 1 to 10 and the number of negative samples $M$ is set to be 1. $\lambda$ is set to be 0.25. In the second training stage, we freeze BERT model and train the decision boundary only. The batch size is set to be 32, $e$ in range 0.5 to 1.2, $s$ from 0 to 0.5, $\eta$ from 0 to 1. We utilize Adam to optimize the model with a learning rate of 2e-5.
\begin{table}
  \centering
    {\renewcommand{\arraystretch}{0.9}
    \resizebox{1.0\columnwidth}{!}{
  \begin{tabular}{cccccc}
    \toprule
    \textbf{Dataset}   & \textbf{Class} & \textbf{Train/Valid/Test}  & \textbf{Length (max/mean)} \\ 
    \midrule
    BANKING & 77 & 9003 / 1000 / 3080 & 79 / 11.91 \\
    OOS & 150 & 15000 / 3000 / 5700 & 28 / 8.31 \\
    StackOverflow & 20 & 12000 / 2000 / 6000 & 41 / 9.18 \\
  \bottomrule
  \end{tabular}}}
\caption{Statistics of experimental datasets.}
\label{tab:commands}
\end{table}

\begin{table*}[]
\centering
    {\renewcommand{\arraystretch}{1.0}
   \resizebox{1.8\columnwidth}{!}{
\begin{tabular}{c|c|ccc|ccc|ccc}
\toprule
 & & \multicolumn{3}{c|}{\textbf{BANKING}} & \multicolumn{3}{c|}{\textbf{OOS}} & \multicolumn{3}{c}{\textbf{StackOverflow}} \\
 & \textbf{Methods} &  \textbf{ALL} & \textbf{Unknown} & \textbf{Known} &  \textbf{ALL} & \textbf{Unknown} & \textbf{Known}  & \textbf{ALL}  & \textbf{Unknown} & \textbf{Known} \\
\midrule
 & MSP  & 50.09 & 41.43 & 50.55  & 47.62 & 50.88 & 47.53  & 37.85 & 13.03 & 42.82 \\ 
 & DOC & 58.03 & 61.42 & 57.85  & 66.37 & 81.98 & 65.96  & 47.73 & 41.25 & 49.02 \\ 
 & OpenMax  & 54.14 & 51.32 & 54.28  & 61.99 & 75.76 & 61.62  & 45.98 & 36.41 & 47.89 \\ 
25\% & DeepUnk & 61.36 & 70.44 & 60.88  & 71.16 & 87.33 & 70.73  & 52.05 & 49.29 & 52.60 \\ 
 & ADB  & 71.62 & 84.56 & 70.94  & 77.19 & 91.84 & 76.80 & 80.83 & 90.88 & 78.82 \\ 
 & CLAB &  \textbf{75.87} & \textbf{88.73} & \textbf{75.20}&  \textbf{81.26} & \textbf{94.49} & \textbf{80.91} & \textbf{86.03} & \textbf{95.12} & \textbf{84.21} \\ \midrule
 & MSP & 71.18 & 41.19 & 71.97 & 70.41 & 57.62 & 70.58  & 63.01 & 23.99 & 66.91 \\ 
 & DOC  & 73.12 & 55.14 & 73.59 & 78.26 & 79.00 & 78.25  & 62.84 & 25.44 & 66.58 \\ 
 & OpenMax  & 74.24 & 54.33 & 74.76 & 80.56 & 81.89 & 80.54 & 68.18 & 45.00 & 70.49 \\ 
50\% & DeepUnk  & 77.53 & 69.53 & 77.74  & 82.16 & 85.85 & 82.11  & 68.01 & 43.01 & 70.51 \\ 
 & ADB & 80.90 & 78.44 & 80.96 & 85.05 & 88.65 & 85.00  & 85.83 & 87.34 & 85.68 \\ 
 & CLAB &  \textbf{83.08} & \textbf{81.82} & \textbf{83.36}  & \textbf{87.03} & \textbf{90.63} & \textbf{86.98} &  \textbf{87.68} & 	\textbf{89.30} & \textbf{87.52} \\ \midrule
 & MSP  & 83.60 & 39.23 & 84.36  & 82.38 & 59.08 & 82.59  & 77.95 & 33.96 & 80.88 \\ 
 & DOC  & 83.34 & 50.60 & 83.91  & 83.59 & 72.87 & 83.69  & 75.06 & 16.76 & 78.95 \\ 
 & OpenMax  & 84.07 & 50.85 & 84.64  & 73.16 & 76.35 & 73.13 & 79.78 & 44.87 & 82.11 \\ 
75\% & DeepUnk  & 84.31 & 58.54 & 84.75 &  86.23 & 81.15 & 86.27  & 78.28 & 37.59 & 81.00 \\ 
 & ADB  & 85.96 & 66.47 & 86.29 & 88.53 & 83.92 & 88.58  & 85.99 & 73.86 & 86.80 \\ 
 & CLAB &  \textbf{88.12} & \textbf{70.74} & \textbf{88.42}  & \textbf{90.53} & \textbf{86.91} & \textbf{90.57} & \textbf{88.11} & \textbf{77.59} & \textbf{88.81} \\
\bottomrule
\end{tabular}}}
\caption{The results of open intent classification by varying the proportions (25\%, 50\% and 75\%) of known classes. Here,  ``ALL'', ``Unknown'' and ``Known'' denote the macro F1 score over all intent classes, unknown (open) intent class and known classes, respectively. }
\label{tab:result_table}
\end{table*}

\section{Experimental Results}
\subsection{Main Results}
The results are shown in the Table \ref{tab:result_table}. We report the macro F1 scores of all compared methods over all intent classes (denoted as ``ALL''), unknown (open) intent class (denoted as ``Unknown'') and known classes (denoted as ``Known''), respectively 
From the results, we can observe that our method substantially and consistently outperforms the state-of-the-art baseline methods by a noticeable margin, which demonstrates the effectiveness of our method. Specifically, compared with the most competitive ADB method, our method improves macro F1 scores by 4.25\%/2.18\%/2.16\% on BANKING, by 4.07\%/1.98\%/2.00\% on OOS , and by 5.20\%/1.85\%/2.12\% on StackOverflow with the proportions of 25\%/50\%/75\% known classes, respectively. 

\begin{table}[]
    {\renewcommand{\arraystretch}{1.0}
    \resizebox{1.0\columnwidth}{!}{
\begin{tabular}{c|c|ccc}
\toprule
\textbf{Dataset}  & \textbf{Method}  & \textbf{ALL} & \textbf{Unknown} & \textbf{Known} \\ \midrule
\multirow{4}{*}{BANKING} & ADB & 71.62 & 84.56 & 70.94  \\ 
 & CLAB &  \textbf{75.87} & \textbf{88.73} & \textbf{75.20} \\
 & w/o ADBES & 74.11  & 86.17 & 73.47 \\ 
 & w/o KCCL  & 73.10  & 86.28 & 72.41 \\ 
 \midrule
\multirow{4}{*}{OOS} & ADB  & 77.19 & 91.84 & 76.80  \\ 
 & CLAB  & \textbf{81.26} & \textbf{94.49} & \textbf{80.91} \\
 & w/o ADBES   & 79.25 & 92.85 & 78.89  \\ 
 & w/o KCCL   & 78.86  & 92.61 & 78.50 \\
 \midrule 
\multirow{4}{*}{StackOverflow} & ADB  & 80.83 & 90.88  & 78.82 \\ 
 & CLAB  & \textbf{86.03} & \textbf{95.12} & \textbf{84.21} \\ 
 & w/o ADBES  & 84.41& 94.15  & 82.47  \\ 
 & w/o KCCL  & 83.35 & 92.88  & 81.44 \\ 
\bottomrule
\end{tabular}}}
\caption{Ablation results of CLAB on three datasets over  with 25\% known classes}
\label{table:ablation}
\end{table}

In addition, we observe that our method not only obtains significant improvements on unknown classes, but also substantially improves the performances on known classes compared with the baselines. This may be because our method can learn accurate and suitable decision boundaries for intent classes.

\subsection{Ablation Study}
To verify the effectiveness of KCCL and ADBES methods for open intent classification, we perform ablation tests on the three datasets with 25\% known classes. In particular, we remove the KCCL (denoted as w/o KCCL) from our proposed CLAB method by replacing the first stage loss function  $\mathcal{L}_{S_1}$ with the cross-entropy loss $\mathcal{L}_{\rm CE}$. In addition,  we remove the ADBES (denoted as w/o ADBES) from CLAB by replacing the decision boundary learning method with ADB \cite{zhang2021deep}.

The ablation test results are summarized in Table \ref{table:ablation}. The performance of CLAB drops sharply when discarding KCCL. This is because KCCL enables CLAB to learn more robust and balanced text representations. In addition, ADBES also makes a great contribution to CLAB. This is reasonable since ADBES plays a critical role in learning adjustable decision boundaries for the known intents. Not surprisingly, both KCCL and ADBES techniques contribute a noticeable improvement to our method, since the performance of ADB (without both KCCL and ADBES) is much worse than that of CLAB. This is within our expectation since the KCCL can learn better feature representations and ADBES can learn better adjustable decision boundaries of known intents.

\begin{table}[]
    {\renewcommand{\arraystretch}{1.0}
    \resizebox{1.0\columnwidth}{!}{
\begin{tabular}{c|c|ccc}
\toprule
\textbf{Percent}  & \textbf{Method} &  \textbf{ALL} & \textbf{Unknown} & \textbf{Known} \\ 
\midrule
 & ADBES & 73.10 & 86.28 & 72.41  \\
25\% & KCL+ADBES  & 74.16 & 86.31 & 73.52 \\
 & KCCL+ADBES &  \textbf{75.87} & \textbf{88.73} & \textbf{75.20} \\\midrule

 & ADBES  & 81.46 & \textbf{81.86} & 81.45 \\
50\% & KCL+ADBES  & 81.95 & 81.12 & 82.19  \\
 & KCCL+ADBES  & \textbf{83.08} & 81.82 & \textbf{83.36} \\ \midrule
   & ADBES  & 86.32  & 69.02 & 86.59 \\
75\% & KCL+ADBES  & 87.12 & 70.14 & 87.40 \\
 & KCCL+ADBES  & \textbf{88.12} & \textbf{70.74} & \textbf{88.42} \\ 
\bottomrule
\end{tabular}}}
\caption{Performance comparison between KCCL and KCL on the BANKING dataset.}
\label{table:loss_method}
\end{table}

\subsection{Analysis of K-center Contrastive Learning}
\paragraph{Performance Comparison between KCCL and KCL}
To investigate the advantage of KCCL over the previously published K-positive contrastive learning (KCL) \cite{kang2020exploring}, we conduct experiments by varying the proportions (25\%, 50\%, and 75\%) of known classes on the BANKING dataset. For a fair comparison, for the KCL, we also use the ADBES method to learn the decision boundary. We run each experiment ten times and report the average results in Table \ref{table:loss_method}. From the results, we can observe that our KCCL method performs significantly better than KCL, especially with 25\% known intents. This verifies the effectiveness of KCCL in learning better relations among the positive samples. 

\begin{figure}
    \centering
    \includegraphics[height=105pt]{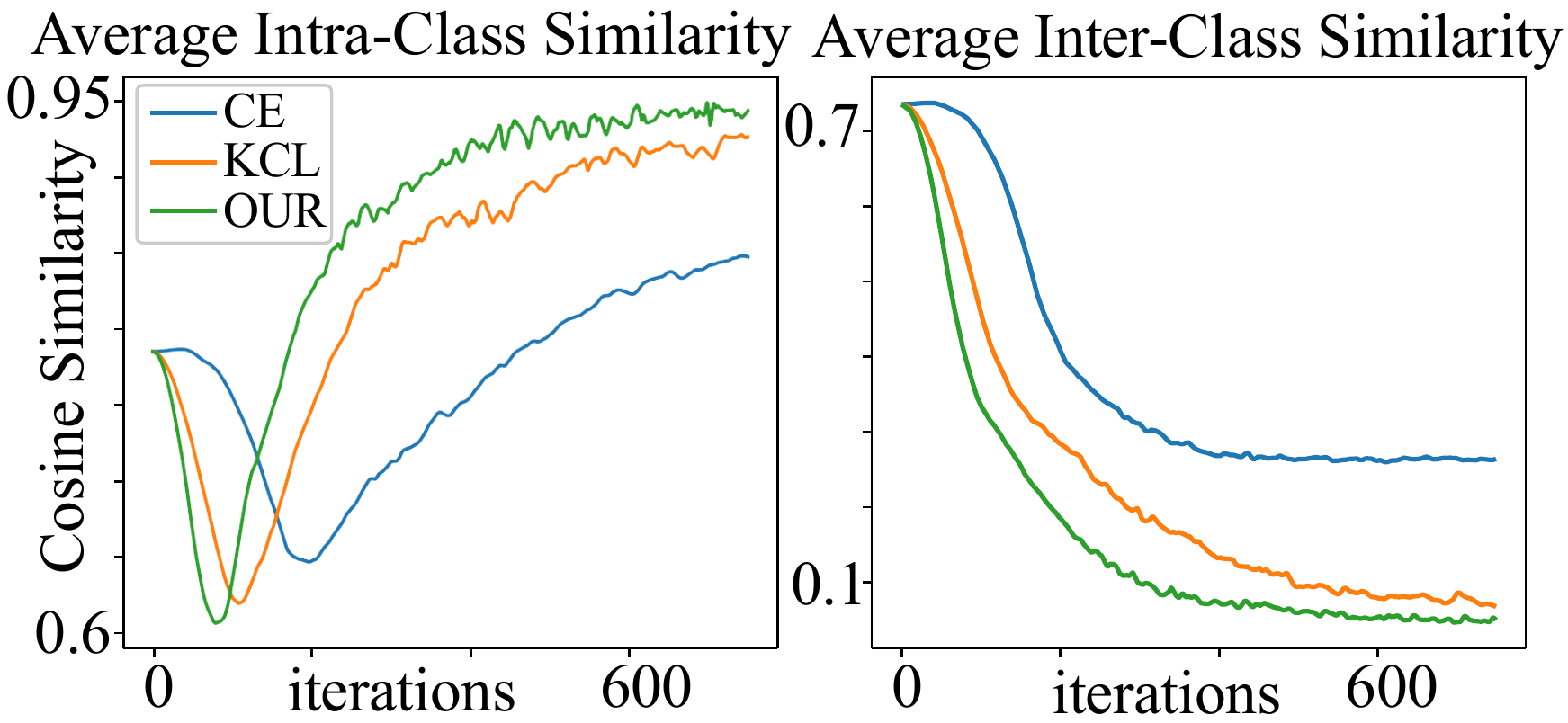}
    \caption{Similarity comparison of different loss functions on BANKING with 25\% known classes. 10 random sentences are sampled from the test set of each class. Figure (a) denotes intra-class cosine similarities. Figure (b) denotes inter-class cosine similarities.}
    \label{fig:loss_image}
\end{figure}

In addition, we also report the learning curves corresponding to the average cosine similarity between intra-class and inter-class samples on the BANKING dataset. As illustrated in Figure \ref{fig:loss_image}, our model can achieve better upper bounds and faster convergence speed  than the compared methods. Specifically, the average similarity between intra-class samples increases sharply until 400 iterations, while the average similarity between inter-class samples gets converged until 400 iterations. This verifies the effectiveness of our KCCL for learning more discriminative representations than the compared methods.   


\paragraph{Effect of Positive Samples in KCCL}
To verify the effect of the number of positive samples ($K$), we conduct experiments by varying the value of $K$ on three datasets with the proportion 25\% of known intents. We report the F1 results in Table \ref{tab:k_compare}. 
We can observe from the results that different values of $K$ have slightly fluctuations on the final results. In addition, we also notice that the optimal $K$ value for different datasets is different. The model with a smaller number of positive samples can achieve the best effect on the three corpora.
Generally, our method is not very sensitive to the number of positive samples (i.e., $K$). 

\begin{table}[]
\centering
\fontsize{8.5}{9}\selectfont
\begin{tabular}{cccc}
\toprule
\textbf{$K$} & \textbf{BANKING} & \textbf{OOS} & \textbf{StackOverflow} \\ \midrule
1 & 72.58 & 80.18 & 86.25 \\
2 & 73.31 & 80.27 & 86.39 \\
3 & 73.21 & \textbf{80.86} & 86.37 \\
4 & 74.83 & 80.58 & \textbf{86.61} \\
5 & \textbf{75.01} & 80.49 & 86.26 \\
6 & 74.55 & 80.64 & 85.99 \\
7 & 72.99 & 80.75 & 85.31 \\
8 & 74.32 & 80.74 & 84.31 \\
9 & 73.01 & 80.34 & 84.60 \\
10 & 72.97 & 80.74 & 85.18 \\
\bottomrule
\end{tabular}
\caption{The macro F1 score of our model by varying the value of $K$ on three datasets with the proportion 25\% of known intents. Here, $K$ denotes the number of positive examples used in KCCL.}
\label{tab:k_compare}
\end{table}

\subsection{Analysis of Adjustable Decision Boundary}
\begin{figure}
    \centering
    \includegraphics[height=145pt]{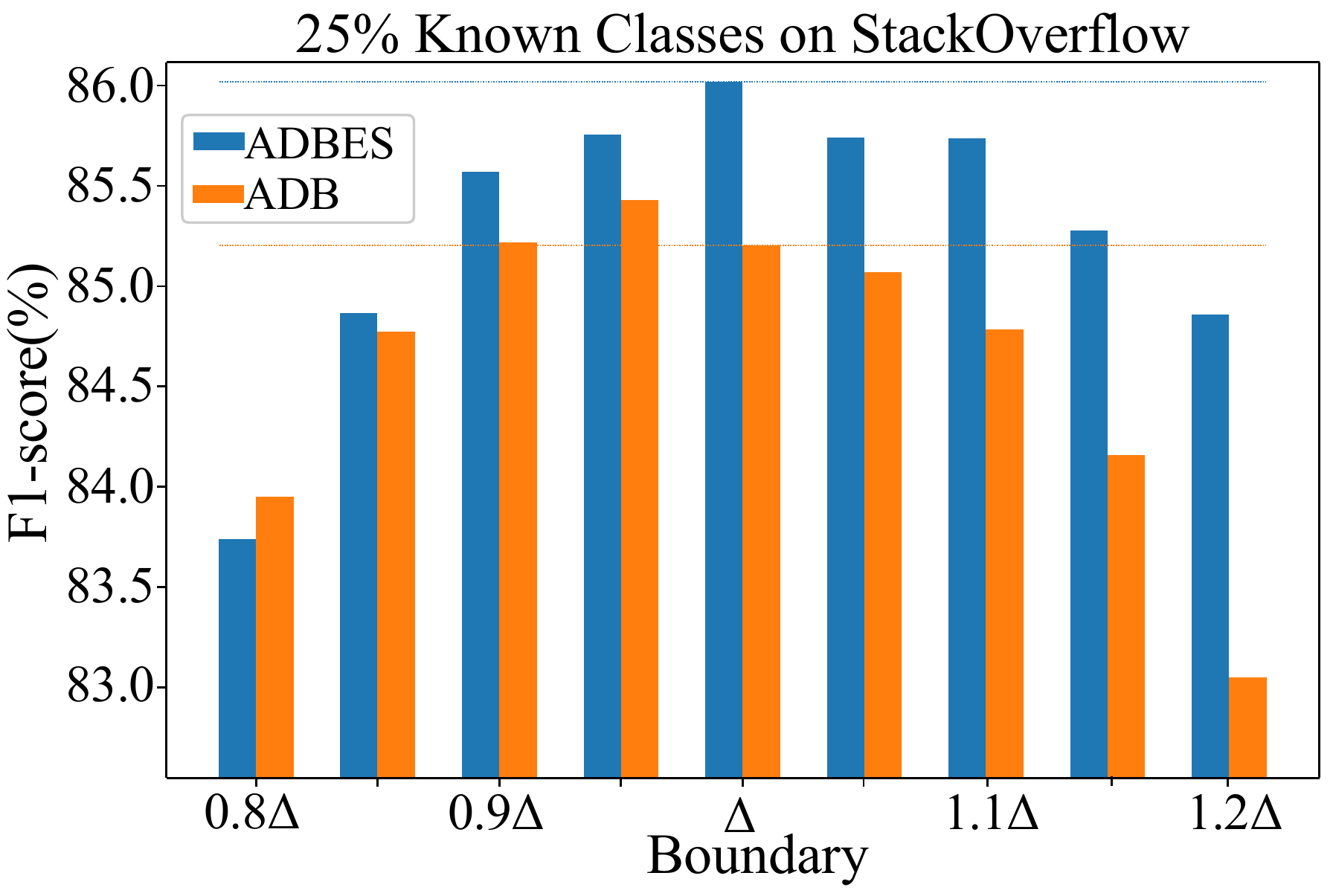}
    \caption{Influence of the decision boundary during inference. The blue and orange dash lines denotes the results learned by ADBES and ADB respectively. }
    \label{fig:boundary_compare}
\end{figure}

To further investigate the tightness and coverage of decision boundary learned by our ADBES method and the ADB method, we use different ratios of $\Delta$ as boundaries during testing. The change of the decision boundary is limited to a small interval, as previous research has found that a large decision boundary change will inevitably lead to a significant drop in the F1-score \cite{zhang2021deep}. In particular, for both ADB and ADBES methods, we incorporate KCCL with the fine-tuned BERT models to learn intent features of known intents, and then the ADBES and ADB methods are applied to learn the spherical decision boundary for each intent class. The experimental results are shown in Figure  \ref{fig:boundary_compare}. We observe that ADBES achieves the best performance with the learned $\Delta$ among all tested decision boundaries, verifying the effectiveness of the decision boundaries learned by ADBES. However, the best result is achieved by ADB when using $0.95\Delta$, which means that the decision boundaries learned by ADB are overrelaxed. 
 

\subsection{Distance Distribution Visualization}

\begin{figure*}[t]
    \centering
    \includegraphics[height=300pt]{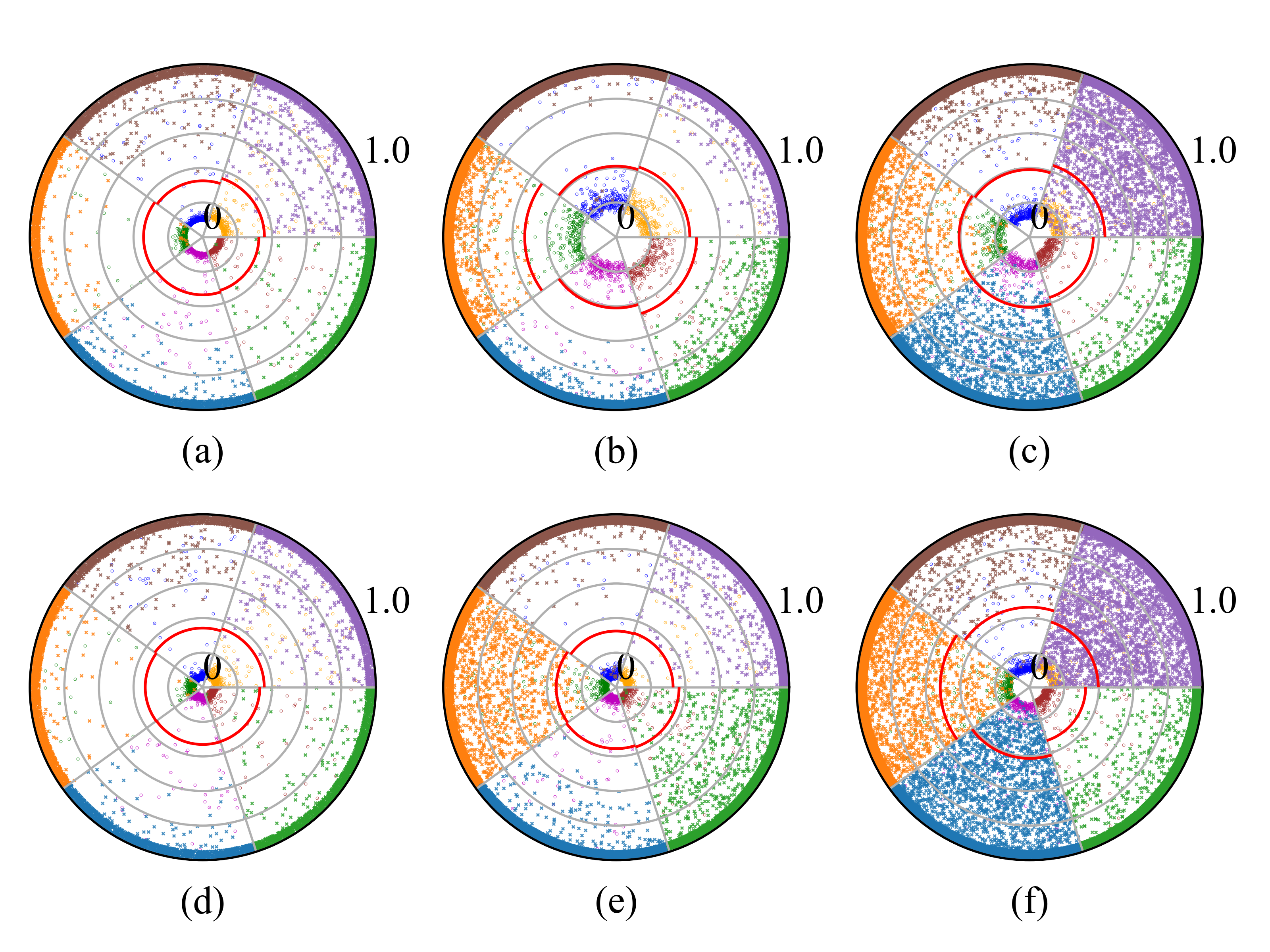}
    \caption{Distance distribution visualization. Figures (a), (b) and (c) represent the distributions and decision boundaries learned by our method with proportions (25\%, 50\%, and 75\%) of known classes on the StackOverflow dataset. Figures (d), (e) and (f) represent the distributions learned by the ADB method. The circles are divided into 5 sectors representing 5 classes. The points close to the center of each circle are known class samples, while the points that have different colors with the known classes represent open class samples. The center of the circle represents the decision center of the corresponding class. The  graphs are drawn according to the Euclidean distance from the sample points to the decision centers. We reduce the value of the distance exceeding 1 to 0.99. We use the red arc to represent the decision boundary.}
    \label{fig:distance_image}
\end{figure*}

We visualize the distance distribution and decision boundary of our CLAB method and ADB (without KCCL training). 
The experiments are conducted on the test set with proportions (25\%, 50\%, and 75\%) of known classes on the StackOverflow dataset. For a fair comparison, we normalize the sentence representations produced by the original ADB method. For each dataset, we randomly selected five categories for demonstration.

As shown in Figure \ref{fig:distance_image}, we have two observations. First, the distance distribution overlap between open intent samples and known intent samples learned by the ADB baseline is less than our CLAB method, especially with 25\% and 50\% known classes. Second, the decision boundaries learned by our CLAB method are more accurate and effective for open intent classification, especially when the proportion of known classes is 25\%. The ability of our method to identify the open intents is significantly stronger than that of the ADB method.


\section{Conclusion}
In this paper, we proposed a novel two-stage method CLAB to improve the effectiveness of open intent classification. 
First, we devised a K-center contrastive learning algorithm to learn discriminative and balanced intent features, improving the generalization of the model for recognizing open intents.
Second, we introduced an adjustable decision boundary learning method with expanding and shrinking to determine the suitable decision conditions. In particular, we expanded the radius of the decision boundary to accommodate more in-class instances if the out-of-class instances were far from the decision boundary; otherwise, we shrunk the radius of the decision boundary.
Extensive experiments on three benchmark datasets 
showed the effectiveness of CLAB. 

\section*{Acknowledgements}
This work was supported by the National Key Research and Development Program of China (No. 2022YFF0902100), the National Natural Science Foundation of China (61906185, 62006062, 62176076), Shenzhen Science and Technology Innovation Program (Grant No. KQTD20190929172835662), Shenzhen Basic Research Foundation (No. JCYJ20210324115614039 and No. JCYJ20200109113441941), Chinese Key-Area Research and Development Program of Guangdong Province (2020B0101350001), and Guangdong Provincial Key Laboratory of Big Data Computing, the Chinese University of Hong Kong, Shenzhen.


\bibliography{aaai23}

\end{document}